\documentclass[runningheads]{llncs}

\usepackage[T1]{fontenc}
\usepackage{graphicx,verbatim}
\usepackage{bm}
\usepackage{amsmath}
\usepackage{amssymb}
\usepackage{multirow}
\usepackage{array}
\usepackage{booktabs}
\usepackage{subfigure}
\usepackage{xcolor}
\usepackage{graphicx}
\usepackage{amssymb}
\usepackage{amsmath}
\usepackage{bm}
\usepackage{pifont}
\usepackage{multirow}
\usepackage{booktabs}
\usepackage{subfigure}
\usepackage{makecell}
\usepackage{xurl}
\usepackage{array}
\usepackage{xcolor}
\usepackage{esvect}
\usepackage[export]{adjustbox}
\usepackage{graphicx}
\usepackage[colorlinks,
            linkcolor=blue,
            anchorcolor=blue,
            citecolor=blue]{hyperref}
\usepackage{graphicx,verbatim}
\usepackage{wrapfig}
\usepackage{marvosym}
\begin{document}
\title{Model Merging for Medical LVLMs: A Benchmark and a Winner-Take-All Approach}
\titlerunning{Model Merging for Medical LVLMs}

\author{Lichao Mou\inst{1} \and
Shilan Zhang\inst{2}\thanks{Work done during an internship at MedAI Technology (Wuxi) Co. Ltd.} \and
Chunlei Li\inst{1} \and
Bingcong Yan\inst{1} \and
Jingliang Hu\inst{1} \and
Yilei Shi \inst{1} \and
Shengwu Xiong\inst{2} \and
Xiao Xiang Zhu\inst{3} \and
Lei Li\inst{4} \and
Yaxiong Chen\inst{2}\textsuperscript{(\Letter)}}

\authorrunning{L. Mou et al.}
\institute{MedAI Technology (Wuxi) Co. Ltd. \and Wuhan University of Technology\\\email{chenyaxiong@whut.edu.cn} \and Technical University of Munich \and National University of Singapore}

\maketitle

\begin{center}
\small
\url{https://github.com/MedAI-T/MergeMedBench}
\end{center}

\begin{abstract}
Large vision-language models (LVLMs) can be adapted to specialized medical imaging tasks via parameter-efficient fine-tuning approaches such as low-rank adaptation (LoRA), leading to a growing ecosystem of expert models tailored to specific imaging modalities and clinical scenarios. However, deploying multiple expert LVLMs in practice incurs substantial computational and operational overhead. Model merging provides a promising solution by consolidating multiple experts into a single model without retraining, yet it remains largely unexplored in the medical domain. In this work, we present the first systematic study of model merging for medical LVLMs. We introduce \textbf{MergeMedBench}, a comprehensive benchmark spanning eight imaging modalities and diverse clinical task types, comprising 16 LoRA fine-tuned models built upon two mainstream architectures. We conduct an extensive evaluation of existing merging methods and further propose winner-take-all, a simple and hyperparameter-free approach that retains only the most dominant parameters across expert models. By preserving the critical parameters that govern model behavior and discarding weaker ones, our method avoids the information dilution inherent in averaging- or alignment-based strategies. Despite its simplicity, winner-take-all consistently outperforms existing approaches, offering both a new perspective on LoRA merging and a strong practical baseline for future research.


\end{abstract}

\section{Introduction}
\label{sec:intro}
Large vision-language models (LVLMs), such as Qwen-VL \cite{ref:Qwen3VL} and InternVL \cite{ref:InternVL}, have become foundational architectures for multimodal understanding. Beyond strong performance on general-purpose tasks, these models can be efficiently adapted to specialized domains through parameter-efficient fine-tuning, most notably low-rank adaptation (LoRA) \cite{ref:LoRA}. This has enabled the development of domain-specific LVLMs, particularly in medical imaging, where fine-tuned models now exist for diverse imaging modalities and clinical scenarios.
\par
However, deploying multiple specialized medical LVLMs introduces substantial computational and operational overhead. This raises a practical question: can we consolidate multiple expert medical LVLMs into a single model that performs well across diverse imaging modalities and clinical tasks? Model merging offers a promising solution by integrating parameters from independently fine-tuned models without retraining.
\par
Various model merging methods have been proposed in natural language processing and computer vision \cite{ref:mmm:IterS,ref:mmm:pcb-m,ref:mmm:RegMean,ref:mmm:LoRAHub,ref:mmm:FisherMerging,ref:mmm:LoRASoups,ref:mmm:ziplora,ref:mmm:LoRArar,ref:mmm:AdaMerging,ref:mmm:ExpertMerging,ref:mmm:Linear,ref:mmm:LoRALEGO}. \cite{ref:TA} introduces task vectors---weight differences between fine-tuned and pre-trained models---and merges them via averaging. However, naive averaging can cause destructive interference when parameter updates conflict across tasks. TIES-Merging \cite{ref:TIES} addresses this by trimming redundant parameters and resolving sign conflicts before merging, while DARE \cite{ref:DARE} reduces redundancy by randomly dropping and rescaling delta parameters. More recently, methods targeting LoRA have emerged: KnOTS \cite{ref:KnOTS} observes that LoRA fine-tuned models exhibit lower weight alignment than fully fine-tuned counterparts and uses singular value decomposition (SVD) to transform LoRA weights into an aligned space; Core Space Merging \cite{ref:Core} constructs a common subspace to improve merging; and RobustMerge \cite{ref:robustmerge} emphasizes directional robustness when combining LoRA modules. Despite these advances, LoRA merging has not been systematically studied for medical LVLMs. Existing methods are developed and evaluated primarily on general language or vision benchmarks, leaving their effectiveness in the medical domain unexplored.
\par
To address this gap, we present the first systematic study of model merging for medical LVLMs. We curate a comprehensive dataset spanning eight imaging modalities across multiple clinical task types. Using this dataset, we fine-tune 16 expert medical LVLMs with LoRA on two widely adopted architectures, Qwen-VL and InternVL, providing a rigorous testbed for evaluating model merging in medical settings.
\par
We also propose a simple yet effective merging method called winner-take-all. Unlike existing approaches that rely on weighted averaging, sign consensus, or subspace alignment, our method follows a different principle: for each position in the LoRA matrices, we select and retain the most dominant parameter across all expert LVLMs. This strategy is motivated by the observation that a small subset of parameters disproportionately governs model behavior and is highly sensitive to modification. Preserving these dominant parameters exactly is crucial for maintaining task performance. In contrast, averaging- and alignment-based methods inevitably modify these critical parameters, causing information loss and performance degradation. Our contributions are:
\begin{itemize}
    \item We present the first systematic study of model merging for medical LVLMs.
    \item We introduce MergeMedBench, a comprehensive benchmark spanning eight imaging modalities and multiple clinical task types, with 16 LoRA fine-tuned models on two mainstream LVLM architectures, to facilitate future research on medical model merging.
    \item We extensively benchmark existing model merging approaches on the proposed benchmark.
    \item We propose a hyperparameter-free LoRA merging method, winner-take-all, which consistently outperforms existing approaches despite its simplicity, offering both a new perspective on LoRA merging and a practical baseline for future research.
\end{itemize}

\section{MergeMedBench}
\subsection{Dataset}
We construct our benchmark using the open-access subset of the OmniMedVQA dataset~\cite{ref:OmniMedVQA}, which contains 88,995 image-question-answer triplets spanning eight medical imaging modalities: CT, MRI, X-ray, ultrasound, dermoscopy, fundus photography, OCT, and microscopy. The dataset covers a wide range of clinically relevant tasks, including modality and anatomical recognition, clinical diagnosis, lesion grading, and biological attribute analysis.
\par
\begin{figure}[t]
\centering
\includegraphics[width=1\textwidth]{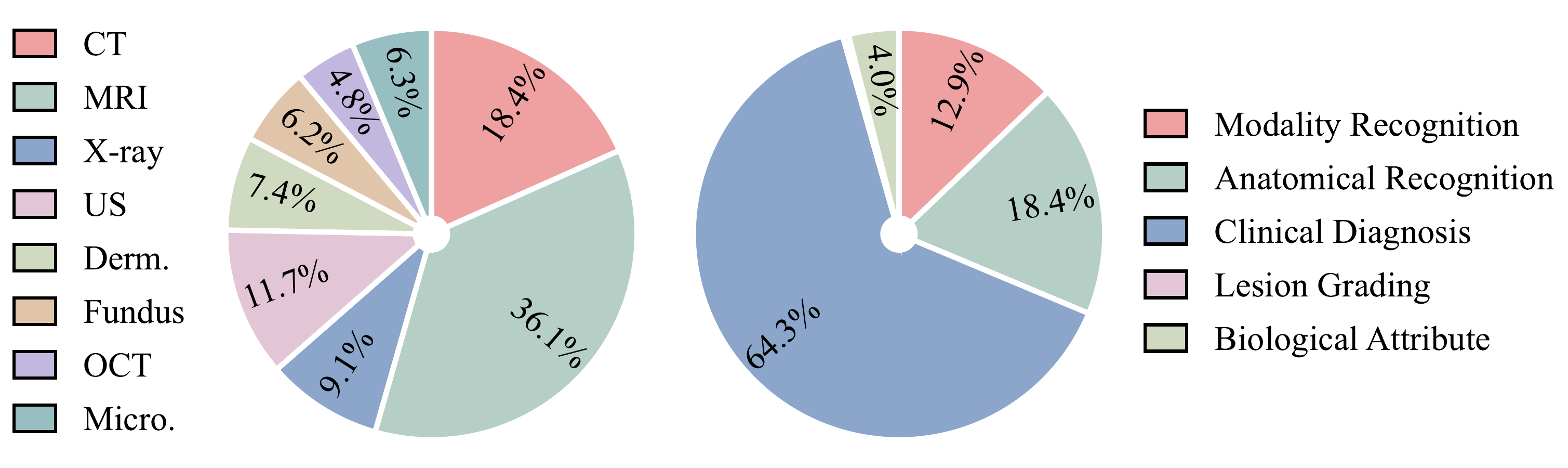}
\caption{Modality (left) and task (right) distributions of the evaluation set.}
\label{fig:benchamrk_distributions}
\end{figure}

We adopt a fixed training/test split to ensure reproducible evaluation and fair comparison. Specifically, 71,333 samples are used to fine-tune pre-trained LVLMs, while the remaining 17,662 samples are reserved for evaluation. Fig.~\ref{fig:benchamrk_distributions} show the modality and task distributions of the evaluation set, respectively. The broad coverage of imaging modalities and task types enables comprehensive assessment of model merging performance in medical vision-language scenarios.

\subsection{Medical LVLMs}
We consider two widely adopted LVLM architectures: Qwen-VL and InternVL. Specifically, we use Qwen3-VL-2B-Instruct and InternVL2-1B as representative models due to their strong performance and widespread adoption in vision-language research, particularly in the medical domain \cite{ref:med1,ref:med2}.
\par
For each architecture, we fine-tune the pre-trained model on individual imaging modalities using LoRA. Following standard practice, LoRA modules are inserted into both attention and feed-forward layers, while base model parameters remain frozen. This yields a collection of expert LVLMs, each optimized for a particular imaging domain and clinical scenario. These expert models serve as inputs to the merging methods evaluated in our benchmark.

\section{Methodology}
\subsection{On the Importance of Dominant Parameters}
\label{sec:observation}

\begin{wrapfigure}[22]{r}{0.5\textwidth}
\vspace*{-2\baselineskip}
\centering
\includegraphics[width=0.48\textwidth]{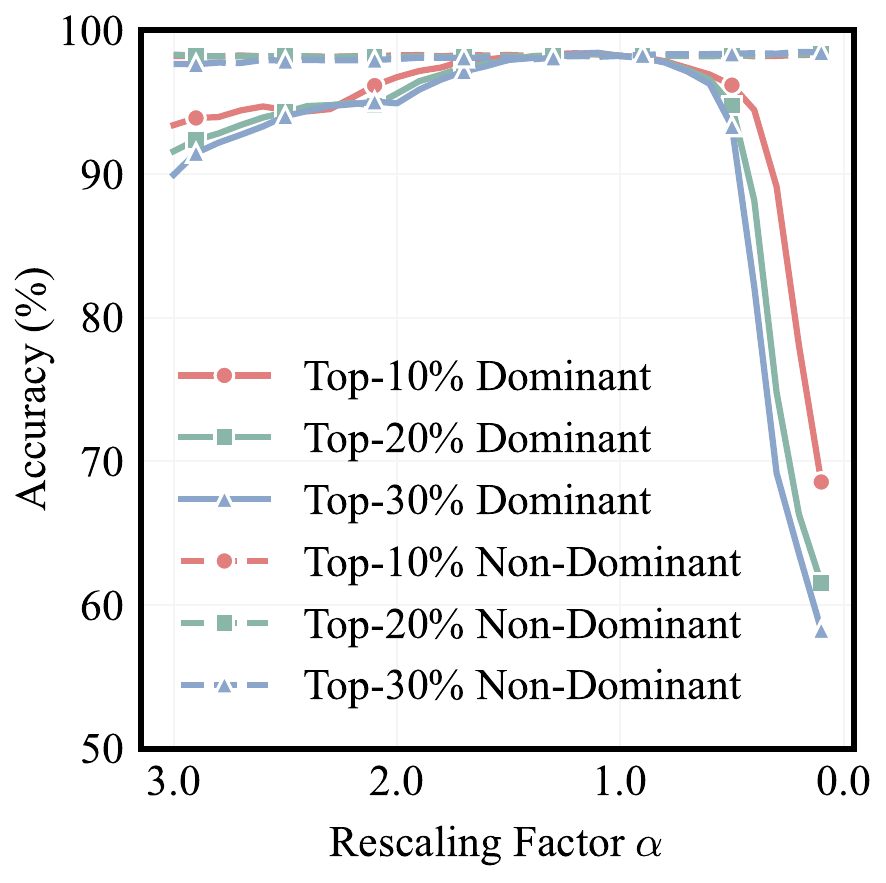}
\caption{Effect of rescaling dominant vs. non-dominant parameters. Dominant parameters are identified using a simple magnitude-based criterion for illustration. This motivates our winner-take-all merging strategy that prioritizes preserving dominant parameters.}
\label{fig:parameters}
\end{wrapfigure}

Before introducing our method, we present a key observation that motivates our approach. Within learned LoRA parameters, a small subset disproportionately governs model behavior; we call these dominant parameters. These parameters encode critical task-relevant information but are highly sensitive to modification. In contrast, non-dominant parameters can be altered substantially with little effect on performance. Consequently, identifying and preserving dominant parameters is essential for effective model merging.
\par
To study this empirically, we conduct a controlled rescaling analysis on LoRA fine-tuned LVLMs. We partition LoRA parameters into dominant and non-dominant subsets based on parameter magnitude\footnote[1]{Magnitude provides a natural proxy for parameter importance; we discuss more general definitions of dominance later.}. We then independently rescale each subset by a factor $\alpha \in \{0, 0.1, \ldots, 1.0, \ldots, 3.0\}$ and measure the resulting performance. As shown in Fig.~\ref{fig:parameters}, rescaling dominant parameters causes dramatic performance shifts---both amplification ($\alpha > 1$) and attenuation ($\alpha < 1$) severely degrade accuracy. In contrast, rescaling non-dominant parameters over the same range produces virtually no change.
\par
This observation has a critical implication for model merging: dominant parameters should be preserved as faithfully as possible. Existing merging methods fail to satisfy this requirement. While approaches such as TIES-Merging~\cite{ref:TIES} explicitly retain high-magnitude parameters, they ultimately rely on averaging to combine contributions from multiple models. This averaging inevitably modifies dominant parameters, causing information loss. The same limitation applies to subspace-based methods~\cite{ref:Core,ref:KnOTS}, which operate in transformed spaces but still employ averaging as the final aggregation step.

\subsection{Problem Formulation}
Consider a pre-trained LVLM with parameters $\theta_0$. Given $N$ downstream tasks, we fine-tune $\theta_0$ on each task independently using LoRA. For each dense layer $l$ and weight matrix $M\in\{W_q, W_k, W_v, W_o\}$, LoRA introduces a low-rank update $\Delta W_{l,M} = B_{l,M} A_{l,M}$, where $B_{l,M} \in \mathbb{R}^{d \times r}$, $A_{l,M} \in \mathbb{R}^{r \times k}$, and the rank $r \ll \min(d, k)$. Let $\{(A_{l,M}^{(n)}, B_{l,M}^{(n)})\}_{n=1}^{N}$ denote the low-rank matrices from fine-tuning on the $N$ tasks. The goal of model merging is to produce a single merged model $\theta_0 + \Delta \theta_{\text{merged}}$ that performs well across all $N$ tasks without additional training, where $\Delta \theta_{\text{merged}}$ denotes all merged weight updates $\{\Delta W_{l,M}^{\text{merged}}\}_{l,M}$.

\subsection{Method}
We propose winner-take-all, a simple yet effective merging strategy grounded in the insight that dominant parameters should remain unmodified. Unlike existing methods that aggregate parameters through weighted averaging, sign consensus, or subspace projection, our approach follows a different principle: for each parameter position in the low-rank matrices, we select and retain the value from the expert that exhibits the strongest dominance according to a scoring function.

\vspace{1ex}
\noindent\textbf{Scoring Function.} We define a scoring function $\mathcal{S}: \mathbb{R} \to \mathbb{R}_{\geq 0}$ to quantify the dominance of individual parameter values. For each low-rank matrix $\Phi\in\{A, B\}$, at each position $(i,j)$ in layer $l$ and weight matrix $M$, the score for the $n$-th expert is:
\begin{equation}
\mathcal{S}_{l,M,i,j}^{(\Phi,n)} = \mathcal{S}\left(\Phi_{l,M}^{(n)}[i,j] \right) \,.
\end{equation}

In principle, $\mathcal{S}$ can be instantiated in various ways. For example, gradient-based scoring such as $\mathcal{S}(w) = |w \cdot \nabla_w \mathcal{L}|$ incorporates task-specific sensitivity information but requires access to task data and incurs additional computation. In this work, we focus on gradient-free merging and use magnitude-based scoring.
\par
A natural choice is $\mathcal{S}(w) = |w|$, reflecting the intuition that larger-magnitude parameters exert stronger influence on model activations. However, directly comparing raw magnitudes across different LoRA fine-tuned LVLMs can be suboptimal. LoRA parameter distributions may differ across experts due to differences in optimization trajectories and training data. Naive magnitude comparison may therefore select parameters that are large due to scale differences rather than true dominance, systematically favoring certain experts.
\par
To address this, we introduce a normalized magnitude-based scoring scheme:
\begin{equation}
\mathcal{S}_{l,M,i,j}^{(\Phi,n)} = 
\left|
\frac{
\Phi_{l,M}^{(n)}[i,j] - \mathbb{E}\big[\Phi^{(n)}\big]
}{
\sqrt{\mathrm{Var}\big[\Phi^{(n)}\big] + \epsilon}
}
\right| \,,
\end{equation}
where $\mathbb{E}[\cdot]$ and $\mathrm{Var}[\cdot]$ denote the mean and variance over all entries in $\Phi^{(n)}$, and $\epsilon$ is a small constant for numerical stability. Normalization ensures robust scoring across experts with different parameter scales.

\vspace{1ex}
\noindent\textbf{Winner Selection.} For each parameter position $(i,j)$ at layer $l$ and weight matrix $M$, we independently select the winning expert for each low-rank matrix $\Phi\in\{A, B\}$:
\begin{equation}
n^{*(\Phi)}_{l,M,i,j} = \underset{n \in \{1, \ldots, N\}}{\arg\max} \; \mathcal{S}_{l,M,i,j}^{(\Phi,n)} \,.
\end{equation}
\par
The merged parameter at each position is then directly taken from the winning expert without rescaling or transformation:
\begin{equation}
\Phi_{l,M}^{\text{merged}}[i,j] = \Phi_{l,M}^{(n^{*(\Phi)}_{l,M,i,j})}[i,j] \,.
\end{equation}

The complete merged LoRA update is:
\begin{equation}
\Delta W_{l,M}^{\text{merged}} = B_{l,M}^{\text{merged}} A_{l,M}^{\text{merged}} \,,
\end{equation}
which is applied to the base model to obtain the final merged model $\theta_0 + \Delta \theta_{\text{merged}}$.

\begin{table}[t]
\caption{Comparison of model merging methods based on Qwen-VL. We report accuracy (\%) across eight imaging modalities (CT, MRI, X-ray, ultrasound, dermoscopy, fundus photography, OCT, and microscopy), along with average accuracy and runtime.}
\label{tab:qwen_comparison}
\renewcommand{\arraystretch}{1.0}
\setlength{\abovecaptionskip}{2pt}
\centering
\resizebox{1.0\linewidth}{!}{
\begin{tabular}{l<{\hspace{0.1cm}}|>{\hspace{0.1cm}}cccccccc<{\hspace{0.1cm}}|>{\hspace{0.1cm}}c<{\hspace{0.1cm}}|>{\hspace{0.1cm}}c}
\toprule
\multicolumn{1}{c|}{\textbf{}} &
  \textbf{CT} &
  \textbf{MRI} &
  \textbf{X-ray} &
  \textbf{US} &
  \textbf{Derm.} &
  \textbf{Fund.} &
  \textbf{OCT} &
  \textbf{Micro.} &
  \textbf{Avg.} &
  \textbf{Time} 
   \\ \midrule
Individual & 98.27 & 99.63 & 95.85 & 99.95 & 94.86 & 97.72 & 98.11 & 98.64 & 97.88 & - \\
\midrule
TA~\cite{ref:TA}              & 56.40 & 69.51 & 82.54 & 79.51 & 77.72 & 87.52 & 68.16 & 79.28 & 75.08          & \textbf{0.05}   \\
TIES-Merging~\cite{ref:TIES}            & 86.45 & 90.08 & 91.21 & 88.28 & 83.00 & 91.71 & 79.95 & 84.59 & 86.91          & 1.20   \\
DARE-TIES~\cite{ref:DARE}       & 87.29 & 92.03 & 85.39 & \textbf{96.05} & 81.93 & 92.62 & 78.89 & 82.25 & 87.06          & 32.72  \\
Iso-C~\cite{ref:ISO}           & 52.48 & 63.96 & 81.30 & 75.36 & 75.19 & 82.33 & 66.16 & 76.04 & 71.60          & 13.52  \\
Iso-CTS~\cite{ref:ISO}         & 52.36 & 64.02 & 81.24 & 75.70 & 75.11 & 81.97 & 65.80 & 75.68 & 71.48          & 170.57 \\
STF~\cite{ref:STF}             & 89.66 & 93.28 & 89.91 & 95.32 & \underline{84.84} & \underline{93.53} & \underline{87.50} & 84.41 & \underline{89.81}          & 64.98  \\
RobustMerge~\cite{ref:robustmerge}     & \underline{90.19} & \textbf{96.39} & 89.78 & \underline{95.47} & 82.08 & 92.71 & 87.26 & 82.34 & 89.53          & 0.54   \\
KnOTS~\cite{ref:KnOTS}           & 63.59 & 79.70 & 86.69 & 84.28 & 80.25 & 90.07 & 76.06 & 81.44 & 80.26          & 272.46 \\
KnOTS-TIES~\cite{ref:KnOTS}      & 87.32 & 91.95 & \underline{91.33} & 89.10 & 84.15 & 92.90 & 85.38 & \underline{85.23} & 88.42          & 272.90 \\
KnOTS-DARE-TIES~\cite{ref:KnOTS} & 89.23 & 93.80 & 88.54 & 92.67 & 83.92 & 92.71 & \textbf{92.22} & 84.41 & 89.69          & 303.85 \\
Core~\cite{ref:Core}            & 56.31 & 69.48 & 82.72 & 79.51 & 78.10 & 87.25 & 67.92 & 79.73 & 75.13          & 650.57 \\
Core-TIES~\cite{ref:Core}       & 65.04 & 79.72 & 86.69 & 86.31 & 79.79 & 90.35 & 78.30 & 80.90 & 80.89          & 644.03 \\
Core-DARE-TIES~\cite{ref:Core}  & 76.92 & 80.58 & 76.41 & 80.28 & 73.28 & 83.33 & 71.11 & 74.86 & 77.10          & 635.65 \\
Winner-Take-All & \textbf{93.80} & \underline{96.30} & \textbf{93.50} & 91.51 & \textbf{85.07} & \textbf{94.17} & 86.08 & \textbf{86.58} & \textbf{90.88} & \underline{0.21}   \\ \bottomrule
\end{tabular}
}
\end{table}

\section{Experiment}
\subsection{Baseline Methods and Metrics}
We focus on gradient-free model merging methods for fair comparison with our winner-take-all approach. Specifically, we consider the following representative baselines: Task Arithmetic (TA)~\cite{ref:TA}, TIES-Merging~\cite{ref:TIES}, DARE~\cite{ref:DARE}, Isotropic Merging~\cite{ref:ISO}, STF~\cite{ref:STF}, RobustMerge~\cite{ref:robustmerge}, KnOTS~\cite{ref:KnOTS}, and Core Space Merging (Core)~\cite{ref:Core}. 
\par
DARE is commonly combined with TIES-Merging, referred to as DARE-TIES. Isotropic Merging has two variants: Iso-C, which operates in a common subspace, and Iso-CTS, which operates in both common and task-specific subspaces. Some subspace-based methods, such as KnOTS and Core, integrate with TIES-Merging or DARE-TIES to further enhance merging performance. We denote these variants by appending the corresponding method names (e.g., KnOTS-TIES and Core-DARE-TIES).
\par
We use normalized per-modality accuracy and average normalized accuracy across all eight modalities to evaluate merging performance. Following prior work, normalized accuracy is defined as the ratio of the merged model's performance on each modality to that of the corresponding individually fine-tuned model. We also report merging time to assess computational efficiency.

\begin{table}[t]
\centering
\caption{Comparison of model merging methods based on InternVL.}
\label{tab:internvl_comparison}
\renewcommand{\arraystretch}{1.0}
\setlength{\abovecaptionskip}{2pt}
\centering
\resizebox{1.0\linewidth}{!}{
\begin{tabular}{l<{\hspace{0.1cm}}|>{\hspace{0.1cm}}cccccccc<{\hspace{0.1cm}}|>{\hspace{0.1cm}}c<{\hspace{0.1cm}}|>{\hspace{0.1cm}}c}
\toprule
\multicolumn{1}{c|}{\textbf{}} &
  \textbf{CT} &
  \textbf{MRI} &
  \textbf{X-ray} &
  \textbf{US} &
  \textbf{Derm.} &
  \textbf{Fund.} &
  \textbf{OCT} &
  \textbf{Micro.} &
  \textbf{Avg.} &
  \textbf{Time} 
   \\ \midrule
Individual & 99.81 & 99.73 & 98.58 & 99.95 & 98.09 & 99.73 & 100.00 & 99.55 & 99.43 & - \\
\midrule
TA~\cite{ref:TA}              & 63.68 & 62.75 & 77.21 & 45.23 & 58.27 & 80.15 & 65.92 & 74.32 & 65.94          & \textbf{0.03}  \\
TIES-Merging~\cite{ref:TIES}            & 84.85 & 80.11 & 88.05 & 58.63 & 70.21 & 89.62 & 77.95 & 84.23 & 79.21          & 0.31  \\
DARE-TIES~\cite{ref:DARE}       & 78.99 & 84.46 & 79.20 & 48.55 & \underline{77.72} & 81.88 & 75.35 & 69.91 & 74.51          & 4.56  \\
Iso-C~\cite{ref:ISO}           & 25.08 & 23.85 & 16.84 & 24.06 & 20.90 & 18.85 & 26.18 & 23.24 & 22.38          & 2.34  \\
Iso-CTS~\cite{ref:ISO}         & 25.58 & 24.80 & 16.78 & 25.22 & 23.28 & 18.76 & 28.18 & 23.96 & 23.32          & 38.21  \\
STF~\cite{ref:STF}             & \textbf{89.42} & 85.75 & \underline{90.09} & \underline{67.70} & 75.65 & \textbf{93.26} & 74.76 & \textbf{86.67} & \underline{82.91}          & 15.65  \\
RobustMerge~\cite{ref:robustmerge}     & 81.33 & \underline{86.31} & 88.36 & 51.69 & 77.11 & 88.34 & 79.13 & 78.56 & 78.85          & 0.26  \\
KnOTS~\cite{ref:KnOTS}           & 63.22 & 62.70 & 77.15 & 45.27 & 58.04 & 80.69 & 66.27 & 74.32 & 65.96          & 42.06  \\
KnOTS-TIES~\cite{ref:KnOTS}      & 82.84 & 82.28 & 87.55 & 67.60 & 68.53 & 88.89 & \textbf{87.50} & \underline{86.40} & 81.45          & 42.51  \\
KnOTS-DARE-TIES~\cite{ref:KnOTS} & 75.78 & 79.31 & 77.15 & 31.00 & 63.32 & 75.96 & 64.39 & 59.37 & 65.78          & 46.51  \\
Core~\cite{ref:Core}            & 63.50 & 62.83 & 77.34 & 44.74 & 58.42 & 80.33 & 65.57 & 74.41 & 65.89          & 103.80  \\
Core-TIES~\cite{ref:Core}       & 74.11 & 69.45 & 82.04 & 57.71 & 63.63 & 86.25 & 76.77 & 81.53 & 73.94          & 115.97  \\
Core-DARE-TIES~\cite{ref:Core}  & 79.14 & 76.19 & 72.01 & \textbf{92.14} & 61.41 & 83.24 & 80.90 & 70.36 & 76.92          & 104.32  \\
Winner-Take-All & \underline{88.71} & \textbf{89.97} & \textbf{91.02} & 62.15 & \textbf{81.09} & \underline{92.26} & \underline{86.79} & 85.14 & \textbf{84.64} & \underline{0.17}  \\ \bottomrule
\end{tabular}
}
\end{table}

\subsection{Results}
Tables~\ref{tab:qwen_comparison} and \ref{tab:internvl_comparison} report the performance of all merging methods on our benchmark. Naive averaging-based TA performs poorly, while TIES-Merging improves results by trimming redundant parameters and resolving sign conflicts before averaging. DARE-TIES uses random dropping instead of deterministic trimming, yielding less stable performance. Among baselines, STF---a subspace method from natural language processing---achieves consistently strong performance, whereas other subspace-based methods (e.g., KnOTS-DARE-TIES, Core-TIES) show unstable behavior.
\par
In comparison, our winner-take-all method achieves the best average accuracy: 90.88\% on Qwen3-VL-2B-Instruct and 84.64\% on InternVL2-1B, consistently outperforming all baselines. Regarding efficiency, subspace-based methods incur substantially higher computational cost, whereas winner-take-all is faster than most baselines while delivering superior performance.

\begin{table}[t]
\centering
\caption{Comparison of winner selection strategies in our framework using Qwen-VL.}
\label{tab:ablation_selection_strategy}
\newcolumntype{C}{>{\hspace{0.1cm}}c<{\hspace{0.1cm}}}
\renewcommand{\arraystretch}{1.0}
\setlength{\abovecaptionskip}{2pt}
\centering
\resizebox{0.62\linewidth}{!}{
\begin{tabular}{C|CCCCCCCC}
\toprule
\textbf{$k$} & 1     & 2     & 3     & 4     & 5     & 6     & 7     & 8     \\ \midrule
\textbf{Avg.}    & \textbf{90.88} & 75.76 & 72.48 & 71.46 & 71.07 & 71.05 & 70.98 & 70.94 \\ \bottomrule
\end{tabular}
}
\end{table}

\begin{table}[t]
  \centering
  \newcolumntype{C}{>{\hspace{0.1cm}}c<{\hspace{0.1cm}}}
  \renewcommand{\arraystretch}{1.0}
  \setlength{\abovecaptionskip}{2pt}
  
  \begin{minipage}[t]{0.48\textwidth}
    \centering
    \caption{Results of combining winner-take-all with subspace-based methods.}
    \label{tab:ablation_wta_subspace}
    \resizebox{\linewidth}{!}{
      \begin{tabular}{C|CC|CC}
      \toprule
      \textbf{}  & Core  & Core-Ours & KnOTS & KnOTS-Ours \\ \midrule
      \textbf{Avg.} & 75.13 & \textbf{89.68}     & 80.26 & \textbf{87.95}      \\ \bottomrule
      \end{tabular}
    }
  \end{minipage}\hfill
  \begin{minipage}[t]{0.48\textwidth}
    \centering
    \caption{Results of different normalization methods in the scoring function.}
    \label{tab:ablation_norm}
    \resizebox{0.78\linewidth}{!}{
      \begin{tabular}{C|CCCC}
      \toprule
      \textbf{} & w/o & min-max & RMS   & Ours  \\ \midrule
      \textbf{Avg.}  & 82.89 & 71.82  & 84.59 & \textbf{84.64} \\ \bottomrule
      \end{tabular}
    }
  \end{minipage}
\end{table}

\begin{figure}[t]
\centering
\includegraphics[width=\textwidth]{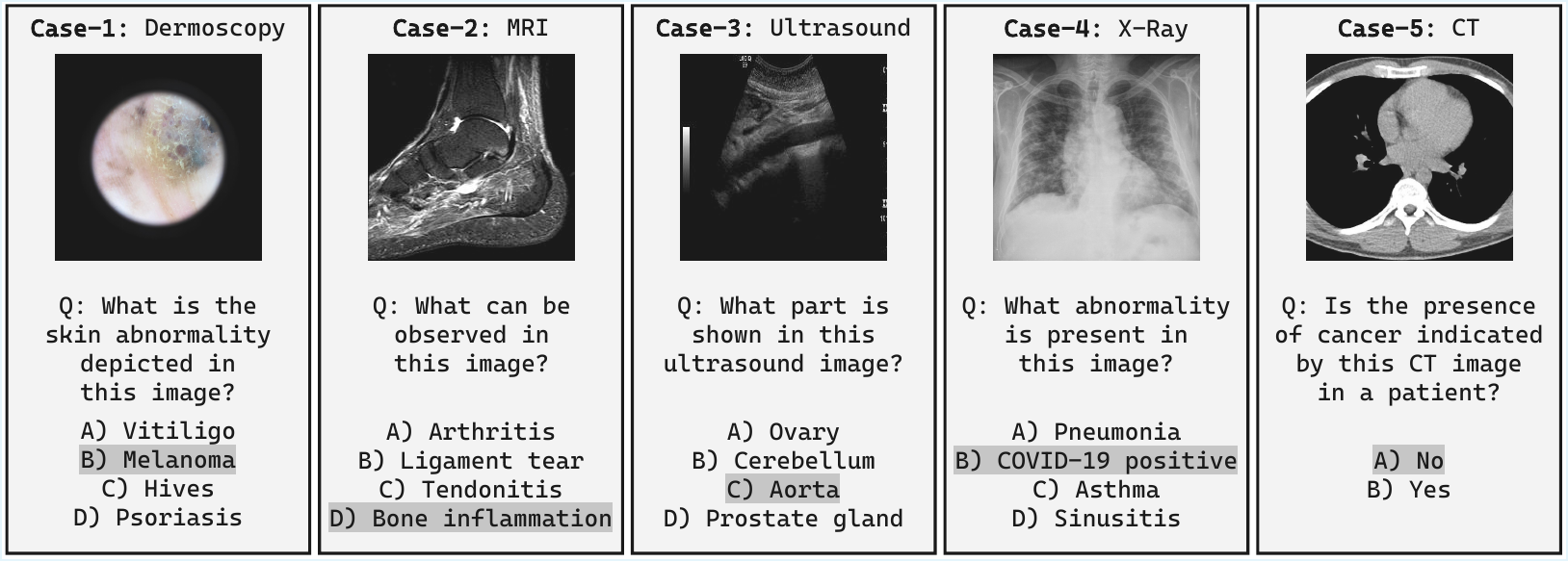}
\caption{Examples of our merged model on multiple modalities and tasks.}
\label{fig:comparison_example}
\end{figure}

\subsection{Discussion}
\noindent\textbf{Winner Selection Strategy.}
Table~\ref{tab:ablation_selection_strategy} evaluates different winner selection strategies, where $k$ denotes selecting the $k$-th ranked expert at each parameter position based on the dominance score. Performance peaks at $k=1$ and decreases as $k$ increases, confirming that preserving the most dominant parameters is crucial.

\vspace{1ex}
\noindent\textbf{Winner-Take-All versus Averaging.}
We compare winner-take-all with simple parameter averaging. Winner-take-all clearly outperforms averaging (90.88\% vs. 72.18\% in average accuracy), supporting our claim that averaging perturbs sensitive dominant parameters, whereas our method preserves them intact.

\vspace{1ex}
\noindent\textbf{Combining Winner-Take-All with Subspace Approaches.}
We integrate winner-take-all with Core and KnOTS and compare the results with their original versions. As shown in Table~\ref{tab:ablation_wta_subspace}, the combined methods improve average accuracy by 14.55\% (Core) and 7.69\% (KnOTS) when merging Qwen-VL fine-tuned models, showing our insight generalizes beyond raw parameter space. Notably, Core with winner-take-all outperforms both Core-TIES and Core-DARE-TIES, while performance remains comparable for KnOTS. These results indicate that winner-take-all is effective both in parameter space and subspace-based merging.

\vspace{1ex}
\noindent\textbf{Effect of Normalization in the Scoring Function.}
Table~\ref{tab:ablation_norm} studies the impact of normalization on InternVL2, where the models exhibit pronounced misalignment. Our normalized magnitude-based scoring improves performance by 1.75\% over the unnormalized variant. We further compare alternative normalization schemes (min-max and RMS), finding our formulation performs best.

\section{Conclusion}
We present the first systematic study of model merging for medical LVLMs. Through extensive evaluation, we benchmark existing merging methods in the medical domain. Furthermore, we propose winner-take-all, a simple yet effective hyperparameter-free strategy that preserves dominant parameters during merging and consistently outperforms prior approaches. We hope this work establishes a strong foundation and spurs future research on efficient merging of medical LVLMs.

\bibliographystyle{splncs04}
\bibliography{refer}

@article{ref:Qwen3VL,
  author       = {Shuai Bai and
                  Yuxuan Cai and
                  Ruizhe Chen and
                  Keqin Chen and
                  Xionghui Chen and
                  Zesen Cheng and
                  Lianghao Deng and
                  Wei Ding and
                  Chang Gao and
                  Chunjiang Ge and
                  Wenbin Ge and
                  Zhifang Guo and
                  Qidong Huang and
                  Jie Huang and
                  Fei Huang and
                  Binyuan Hui and
                  Shutong Jiang and
                  Zhaohai Li and
                  Mingsheng Li and
                  Mei Li and
                  Kaixin Li and
                  Zicheng Lin and
                  Junyang Lin and
                  Xuejing Liu and
                  Jiawei Liu and
                  Chenglong Liu and
                  Yang Liu and
                  Dayiheng Liu and
                  Shixuan Liu and
                  Dunjie Lu and
                  Ruilin Luo and
                  Chenxu Lv and
                  Rui Men and
                  Lingchen Meng and
                  Xuancheng Ren and
                  Xingzhang Ren and
                  Sibo Song and
                  Yuchong Sun and
                  Jun Tang and
                  Jianhong Tu and
                  Jianqiang Wan and
                  Peng Wang and
                  Pengfei Wang and
                  Qiuyue Wang and
                  Yuxuan Wang and
                  Tianbao Xie and
                  Yiheng Xu and
                  Haiyang Xu and
                  Jin Xu and
                  Zhibo Yang and
                  Mingkun Yang and
                  Jianxin Yang and
                  An Yang and
                  Bowen Yu and
                  Fei Zhang and
                  Hang Zhang and
                  Xi Zhang and
                  Bo Zheng and
                  Humen Zhong and
                  Jingren Zhou and
                  Fan Zhou and
                  Jing Zhou and
                  Yuanzhi Zhu and
                  Ke Zhu},
  title        = {{Qwen3-VL} Technical Report},
  journal      = {arXiv preprint arXiv:2511.21631},
  year         = {2025}
}

@article{ref:InternVL,
  author       = {Zhe Chen and
                  Jiannan Wu and
                  Wenhai Wang and
                  Weijie Su and
                  Guo Chen and
                  Sen Xing and
                  Muyan Zhong and
                  Qinglong Zhang and
                  Xizhou Zhu and
                  Lewei Lu and
                  Bin Li and
                  Ping Luo and
                  Tong Lu and
                  Yu Qiao and
                  Jifeng Dai},
  title        = {{InternVL}: Scaling up Vision Foundation Models and Aligning for Generic
                  Visual-Linguistic Tasks},
  journal      = {arXiv preprint arXiv:2312.14238},
  year         = {2023}
}

@inproceedings{ref:LoRA,
  author       = {Edward J. Hu and
                  Yelong Shen and
                  Phillip Wallis and
                  Zeyuan Allen{-}Zhu and
                  Yuanzhi Li and
                  Shean Wang and
                  Lu Wang and
                  Weizhu Chen},
  title        = {{LoRA}: Low-Rank Adaptation of Large Language Models},
  booktitle    = {{International Conference on Learning Representations},},
  year         = {2022}
}

@inproceedings{ref:TIES,
  author       = {Prateek Yadav and
                  Derek Tam and
                  Leshem Choshen and
                  Colin A. Raffel and
                  Mohit Bansal},
  title        = {{TIES-Merging}: Resolving Interference When Merging Models},
  booktitle    = {Advances in Neural Information Processing Systems},
  pages = {7093--7115.},
  year         = {2023}
}

@inproceedings{ref:DARE,
  author       = {Le Yu and
                  Bowen Yu and
                  Haiyang Yu and
                  Fei Huang and
                  Yongbin Li},
  title        = {Language Models are Super Mario: Absorbing Abilities from Homologous Models as a Free Lunch},
  booktitle    = {International Conference on Machine Learning},

  pages        = {57755--57775.},
  year         = {2024}
}

@inproceedings{ref:KnOTS,
  author       = {George Stoica and
                  Pratik Ramesh and
                  Boglarka Ecsedi and
                  Leshem Choshen and
                  Judy Hoffman},
  title        = {Model merging with {SVD} to tie the Knots},
  booktitle    = {{International Conference on Learning Representations},},
  year         = {2025}
}

@article{ref:Core,
  author       = {Aniello Panariello and
                  Daniel Marczak and
                  Simone Magistri and
                  Angelo Porrello and
                  Bartlomiej Twardowski and
                  Andrew D. Bagdanov and
                  Simone Calderara and
                  Joost van de Weijer},
  title        = {Accurate and Efficient Low-Rank Model Merging in Core Space},
  journal      = {arXiv preprint arXiv:2509.17786},
  year         = {2025}
}

@article{ref:robustmerge,
  title={{RobustMerge}: Parameter-efficient model merging for {mllms} with direction robustness},
  author={Zeng, Fanhu and Guo, Haiyang and Zhu, Fei and Shen, Li and Tang, Hao},
  journal={arXiv preprint arXiv:2502.17159},
  year={2025}
}

@inproceedings{ref:OmniMedVQA,
  author       = {Yutao Hu and
                  Tianbin Li and
                  Quanfeng Lu and
                  Wenqi Shao and
                  Junjun He and
                  Yu Qiao and
                  Ping Luo},
  title        = {{OmniMedVQA}: {A} New Large-Scale Comprehensive Evaluation Benchmark
                  for Medical {LVLM}},
booktitle    = {{IEEE/CVF} Conference on Computer Vision and Pattern Recognition},
  pages        = {22170--22183.},
  year         = {2024}
}

@inproceedings{ref:ISO,
  author       = {Daniel Marczak and
                  Simone Magistri and
                  Sebastian Cygert and
                  Bartlomiej Twardowski and
                  Andrew D. Bagdanov and
                  Joost van de Weijer},
  title        = {No Task Left Behind: Isotropic Model Merging with Common and Task-Specific
                  Subspaces},
  booktitle    = {International Conference on Machine Learning,},
  year         = {2025}
}

@inproceedings{ref:STF,
  author       = {Haiquan Qiu and
                  You Wu and
                  Dong Li and
                  Jianmin Guo and
                  Quanming Yao},
  title        = {Superpose Task-specific Features for Model Merging},
  booktitle    = {Empirical Methods in Natural Language Processing},
  pages        = {4200--4214.},

  year         = {2025}
}

@inproceedings{ref:TA,
  author       = {Gabriel Ilharco and
                  Marco T{\'{u}}lio Ribeiro and
                  Mitchell Wortsman and
                  Ludwig Schmidt and
                  Hannaneh Hajishirzi and
                  Ali Farhadi},
  title        = {Editing models with task arithmetic},
  booktitle    = {{International Conference on Learning Representations},},
  year         = {2023}
}

@inproceedings{ref:mmm:AdaMerging,
  author       = {Enneng Yang and
                  Zhenyi Wang and
                  Li Shen and
                  Shiwei Liu and
                  Guibing Guo and
                  Xingwei Wang and
                  Dacheng Tao},
  title        = {{AdaMerging}: Adaptive Model Merging for Multi-Task Learning},
  booktitle    = {{International Conference on Learning Representations},},
  year         = {2024}
}

@article{ref:mmm:ExpertMerging,
  author       = {Dengming Zhang and
                  Xiaowen Ma and
                  Zhenliang Ni and
                  Zhenkai Wu and
                  Han Shu and
                  Xin Jiang and
                  Xinghao Chen},
  title        = {{Expert} Merging: Model Merging with Unsupervised Expert Alignment and
                  Importance-Guided Layer Chunking},
  journal      = {arXiv preprint arXiv:2509.25712},
  year         = {2025}
}

@inproceedings{ref:mmm:Linear,
  author       = {Jinghan Zhang and
                  Shiqi Chen and
                  Junteng Liu and
                  Junxian He},
  title        = {Composing Parameter-Efficient Modules with Arithmetic Operation},
  booktitle    = {Advances in Neural Information Processing Systems},
  pages = {12589--12610.},
  year         = {2023}
}

@inproceedings{ref:mmm:LoRASoups,
  author       = {Akshara Prabhakar and
                  Yuanzhi Li and
                  Karthik Narasimhan and
                  Sham M. Kakade and
                  Eran Malach and
                  Samy Jelassi},
  title        = {{LoRA} Soups: Merging {LoRAs} for Practical Skill Composition Tasks},
  booktitle    = {International Conference on Computational Linguistics},
  pages        = {644--655.},

  year         = {2025}
}

@article{ref:mmm:LoRAHub,
  title={{LoraHub}: Efficient cross-task generalization via dynamic {LoRA} composition},
  author={Huang, Chengsong and Liu, Qian and Lin, Bill Yuchen and Pang, Tianyu and Du, Chao and Lin, Min},
  journal={arXiv preprint arXiv:2307.13269},
  year={2023}
}

@inproceedings{ref:mmm:LoRALEGO,
  author       = {Ziyu Zhao and
                  Tao Shen and
                  Didi Zhu and
                  Zexi Li and
                  Jing Su and
                  Xuwu Wang and
                  Fei Wu},
  title        = {Merging {LoRAs} like Playing {LEGO}: Pushing the Modularity of {LoRA}
                  to Extremes Through Rank-Wise Clustering},
  booktitle    = {{International Conference on Learning Representations},},
  year         = {2025}
}

@inproceedings{ref:mmm:LoRArar,
  title={{LoRA.rar}: Learning to merge {LoRAs} via hypernetworks for subject-style conditioned image generation},
  author={Shenaj, Donald and Bohdal, Ondrej and Ozay, Mete and Zanuttigh, Pietro and Michieli, Umberto},
  booktitle={{IEEE/CVF} Conference on Computer Vision and Pattern Recognition},
  pages={16132--16142.},
  year={2025}
}

@inproceedings{ref:mmm:FisherMerging,
  author       = {Michael Matena and
                  Colin Raffel},
  title        = {Merging Models with Fisher-Weighted Averaging},
  booktitle    = {Advances in Neural Information Processing Systems},
  pages = {17703-17716.},
  year         = {2022}
}

@inproceedings{ref:mmm:pcb-m,
  author       = {Guodong Du and
                  Junlin Lee and
                  Jing Li and
                  Runhua Jiang and
                  Yifei Guo and
                  Shuyang Yu and
                  Hanting Liu and
                  Sim Kuan Goh and
                  Ho{-}Kin Tang and
                  Daojing He and
                  Min Zhang},
  title        = {Parameter Competition Balancing for Model Merging},
  booktitle    = {Advances in Neural Information Processing Systems},
  pages = {84746--84776.},
  year         = {2024}
}

@inproceedings{ref:mmm:IterS,
  author       = {Hongxu Chen and
                  Zhen Wang and
                  Runshi Li and
                  Bowei Zhu and
                  Long Chen},
  title        = {{IterIS}: Iterative Inference-Solving Alignment for {LoRA} Merging},
booktitle    = {{IEEE/CVF} Conference on Computer Vision and Pattern Recognition},
  pages        = {4829--4838.},

  year         = {2025}
}

@inproceedings{ref:mmm:RegMean,
  author       = {Xisen Jin and
                  Xiang Ren and
                  Daniel Preotiuc{-}Pietro and
                  Pengxiang Cheng},
  title        = {Dataless Knowledge Fusion by Merging Weights of Language Models},
  booktitle    = {{International Conference on Learning Representations},},
  year         = {2023}
}

@inproceedings{ref:mmm:ziplora,
  title={{ZipLoRA}: Any subject in any style by effectively merging {LoRAs}},
  author={Shah, Viraj and Ruiz, Nataniel and Cole, Forrester and Lu, Erika and Lazebnik, Svetlana and Li, Yuanzhen and Jampani, Varun},
  booktitle={European Conference on Computer Vision},
  pages={422--438.},
  year={2024},
}

@article{ref:med1,
  author={Lai, Yuxiang and Zhong, Jike and Li, Ming and Zhao, Shitian and Li, Yuheng and Psounis, Konstantinos and Yang, Xiaofeng},
  journal={IEEE Transactions on Medical Imaging}, 
  title={{Med-R1}: Reinforcement Learning for Generalizable Medical Reasoning in Vision-Language Models}, 
  year={2026},
  pages={1-1},
}

@inproceedings{ref:med2,
  author       = {Jiazhen Pan and
                  Che Liu and
                  Junde Wu and
                  Fenglin Liu and
                  Jiayuan Zhu and
                  Hongwei Bran Li and
                  Chen Chen and
                  Cheng Ouyang and
                  Daniel Rueckert},
  title        = {{MedVLM-R1}: Incentivizing Medical Reasoning Capability of Vision-Language
                  Models ({VLMs}) via Reinforcement Learning},
  booktitle    = {Medical Image Computing and Computer Assisted Intervention},

  pages        = {337--347.},
  year         = {2025}
}
\end{document}